# Domain-specific Pretraining Profile and Transformer Performance: Evidence from Modeling Digital Pragmatics in Arabic-English Code-switching

Fahad Al Hussen, King Saud University, Riyadh, Saudi Arabia
Mohammed Q. Shormani, Ibb University, Ibb, Yemen
shormani@ibbuniv.edu.ye/https://orcid.org/0000-0002-0138-4793

(June 6, 2026)

**Abstract**

This study highlights the role of domain-specific pretraining profile (DSPP) in Transformer performance for modeling digital pragmatics in Arabic–English code-switched discourse. It evaluates MARBERT and XLM-R(oBERTa), with BERT serving as a general-purpose baseline. The models were evaluated on their ability to classify context-sensitive pragmatic functions in code-switched social-media discourse. 11695 unique X posts were collected via Python and utilized for the study. The study employs a quantitative and qualitative NLP approach, following a supervised pipeline. Findings unveil that MARBERT consistently surpasses XLM-R with validation Macro F1 increasing from 0.39 to 0.84 and validation loss decreasing from 0.55 to 0.19. On an independent test set, it achieved 0.96 accuracy, 0.83 macro precision, 0.87 macro recall, and 0.85 Macro F1, while XLM-R achieved 0.92 test accuracy but a substantially lower Macro F1 of 0.52. This was also supported by class-level performance where MARBERT outperforms XLM-R considerably with F1 improvements ranging from +0.33 to +0.60, demonstrating a clear advantage in modeling Arabic digital pragmatics. The study concludes that Transformer performance depends more on DSPP than multilingual coverage alone, as the latter does not guarantee optimal performance on a highly specialized pragmatic classification task.

**Keywords**: Domain-specific pretraining profile, Transformers, BERT/MARBERT/XLM-R, Arabic-English code-switching, emoji, digital pragmatics, (self-) attention

## 1. Introduction

Transformer-based models (TBMs) are pretrained language models (PLMs) [36] that have substantially advanced NLP, but their effectiveness in mixed-language and code-switching (CS) environments may depend on how closely their pretraining profile matches the linguistic and communicative characteristics of the target data, which remains underinvestigated. Several questions arise regarding this issue including: What affects PLMs' performance in CS context?, How about multilingual coverage?, and To what extent does Domain-specific Pretraining Profile (DSPP) influence Trasnformer performance? This study addresses DSPP in Arabic-English CS in social media, where nonstandard Arabic dialects (NAD) orthography, bilingual scripts, and English expressions may occur within the same post. It is also relevant to digital/emoji pragmatics because emojis function not simply as decorative or affective symbols but as context-sensitive paralinguistic and pragmatic markers that can signal communicative intent [31], manage discourse, intensify meaning, and contribute to irony or sarcasm [39]. In CS digital discourse, the relationship between text and emojis can extend across languages and scripts, making contextual representation particularly important (cf. [23]).

Although traditional computational approaches have often treated emojis as isolated lexical or affective units [44], their pragmatic functions may emerge from their relationships with surrounding textual material (cf. [38]). This study extends this to their pragmatic functions amalgamated with the linguistic communicative context. Transformer-based PLMs offer powerful contextual representations for such tasks, but their performance may vary according to the linguistic resources and DSPP ([cf. 37]). Thus, an important question is whether pretraining specialization and domain alignment influence performance in Arabic-English CS discourse. This study demarcates whether such pretraining differences are reflected in the classification of emoji pragmatic functions in this CS context. A curated dataset of 11703 unique tweets manually annotated into 9 pragmatic emoji classes, we evaluate 2 models, MARBERT and XLM-R(oBERTA), with BERT serving as general-purpose baseline, under the same training conditions and hyperparameters. BERT tokenizer is not optimized for Arabic, and Arabic forms that are poorly represented in its vocabulary may be mapped to the unknown token ([UNK]), providing a useful baseline for examining the challenges of processing Arabic within the CS data [9]. XLM-R represents massively multilingual pretraining and provides broad cross-lingual coverage [2]. MARBERT, however, is specialized for Arabic and particularly relevant to social-media language, specifically X posts, providing representations more closely aligned with the linguistic domain of our data (cf. [1]). These models therefore allow us to examine not merely a monolingual-versus-multilingual distinction, but the relative importance of general, multilingual, and DSPP.

Thus, this article is set up as follows. Section 2 outlines the theoretical foundations and previous studies and introduces the 3 models used in this study. Section 3 presents the experimental design of the study and pipeline. Section 4 spells out the results, including an evaluation of standard macro-metrics, and an exploration of the mathematical signatures and loss trajectories defining model performance. Section 5 interprets these findings through the lens of computation linguistics, analyzing how tokenization bottlenecks and script blindness impact model inference. Section 6 concludes the paper with limitations, and directions for future research.

## 2. Theoretical foundations and previous studies

### 2.1. Transformer-based models

Transformer-based PLMs represent a major paradigm shift in NLP, introduced by Vaswani et al. [36]. A Transformer could be defined as "a deep learning architecture built around the attention mechanism" processing language by analyzing entire sequences of text simultaneously, but not word-by-word [18, p. 4]. Its core feature is a "self-attention" mechanism, which dynamically calculates mathematical relationships between all words in a sentence to capture context and long-range dependencies regardless of distance. By eliminating the need for sequential processing, it allows for massive parallel computing, making it highly efficient to train on vast datasets (see e.g., [18]). This ability to naturally weigh the importance of different words in context makes it exceptionally skilled at learning complex linguistic nuances, syntax, semantics and cross-lingual patterns. Mathematically, attention is computed using query, key, and value matrices, enabling the model to learn contextual representations dynamically. It follows that Transformer architecture serves as the foundational backbone for modern LLMs that analyze intricate language behaviors such as bilingual CS and emoji usage. Transformer models are composed of stacked encoder and decoder layers; however, modern NLP applications often use encoder-only architectures like BERT or encoder-decoder variants depending on the task. Encoder-only models are particularly effective for classification tasks such as sentiment analysis and pragmatic function detection, which are central to this study (see also [36], [9], [26].

#### 2.1.1. BERT

BERT was first introduced by Devlin et al. [9] designed to pretrain deep contextual representations of text. Unlike traditional unidirectional models, BERT learns context bidirectionally, i.e. from both left and right sides of a token simultaneously, enabling richer semantic understanding. Having 110 million parameters, BERT was pretrained on 2 large English-language text corpora, namely the BookCorpus which contains approximately 800 million words from around 11000 unpublished books, and the English Wikipedia, which contains about 2.5 billion words of text [33]. Its training uses 2 unsupervised tasks, namely masked language modeling (MLM), where random tokens are masked and predicted based on context, and Next Sentence Prediction, which models inter-sentential coherence. This pretraining strategy allows BERT to develop strong general-purpose language representations that can be finetuned for downstream tasks such as text classification, named entity recognition, and sentiment analysis.

However its strengths, BERT has limitations in handling non-English languages and informal social-media text, particularly in morphologically rich and dialectally diverse languages such as Arabic. In this study, BERT is used only as a general-purpose baseline. Although it was not specifically designed for Arabic or code-mixed data, it provides a reference point against which the performance of the principal models, MARBERT and XLM-R, can be contextualized in the Arabic–English CS environment, including socially rich discourse on X and the use of emojis.

#### 2.1.2. MARBERT

MARBERT is another monolingual (bidirectional) TBM developed by [1]; it is a model designed specifically for NAD and social media text. It is, in other words, trained on a large-scale corpus of Arabic tweets, capturing dialectal variation, informal language, and social media-specific expressions. As for its architectural mechanism, MARBERT is based on BERT architecture but optimized for Arabic linguistic diversity. It has "12-layered transformers, and 163 million trainable parameters" [32[, [2]). It is reported that MARBERT is more successful than multilingual models such as AraGPT2 [4], and CAMeLBERT [2], and XLM-R Large [7]. Unlike BERT, MARBERT was trained on large-scale social media outlets' data [11]. In its training process, the MLM objective was used for a large dataset of Arabic X data, making it highly effective for tasks involving sentiment analysis, dialect identification, and social media text classification. Although it is a valuable LLM, it is still limited to Arabic language (see e.g., 32]).

MARBERT is particularly suited to our study because its pretraining is specialized for Arabic social-media language, including NAD and informal orthographic variation [1]. Such specialization may provide richer representations of the Arabic-English code-switching and pragmatic patterns found in X discourse, including emoji pragmatics (cf. [32], [11]. More importantly, its exposure to social-media data may also include English vocabulary and script, making MARBERT potentially well suited to CS Arabic-English contexts. Thus, its performance provides evidence for examining whether DSPP offers an advantage over general-purpose and broadly multilingual pretraining.

#### 2.1.3. XLM-R

XLM-R was developed by Conneau et al. [7] as a large-scale multilingual Transformer-based model trained on 100 languages using a massive CommonCrawl dataset. It extends the R architecture to multilingual settings and removes the need for language-specific models by learning universal cross-lingual representations. What distinguishes XLM-R from earlier multilingual models is that it demonstrates strong transfer learning capabilities across languages, including low-resource languages ([20], [22]. It was trained using an MLM objective similar to BERT but with significantly larger and more diverse training data. XLM-R is particularly effective in multilingual and code-mixed environments, making it suitable for tasks involving both Arabic and English text (cf. [19]. Its shared embedding space allows it to model semantic similarities across languages, which is essential for cross-linguistic pragmatic analysis. In this study, XLM-R is used as the primary multilingual model, enabling comparison with both monolingual English, i.e. BERT, and Arabic-specific MARBERT architectures in the classification of emoji-based pragmatic functions [26].

XLM-R is included as a multilingual pretrained model, trained on text from more than 100 languages, including Arabic, and designed to support cross-lingual representation learning [2], [25]. In this study, it provides a multilingual comparison point alongside the general-purpose English-pretrained BERT and the Arabic social-media-specialized MARBERT. We examine whether its broad linguistic coverage translates into better performance on emoji-pragmatic classification in Arabic-English CS. Its performance thus helps assess the relative contribution of multilingual DSPP contexts.

### 2.2. Code-switching: Theory and practice

Code-switching refers to the alternation of 2 or more languages within a conversation, sentence, or discourse event by bilingual speakers. It represents linguistic deficiency or interference, but it is also a systematic and meaningful communicative strategy governed by linguistic constraints and social motivations [5], 21]. In Arabic–English communication, bilingual speakers frequently switch languages to achieve specific discourse functions such as emphasizing information, expressing emotions, quoting others, signaling expertise, managing interpersonal relationships, or marking shifts in topic and stance (cf. [3]), [45]. CS thus reflects both linguistic competence and communicative intent, making it a central feature of bilingual digital interaction [45].

Several theoretical approaches attempt to explain how and why CS occurs (see e.g., [5], [24], [21]). Structural models emphasize grammatical constraints governing language alternation, while sociolinguistic and interactional approaches focus on the communicative functions served by switching between languages [35]. Within Arabic–English discourse, language alternation often reflects distinctions between local and global identities, prestige, personal and professional domains, or informal and formal communication styles, and sometimes snobbery/showoff. For example, Arabic speakers may use Arabic to express culturally grounded emotions or interpersonal closeness while switching to English for technical terminology, academic concepts, or globally oriented discussions, and sometimes to show off. These language shifts create distinct contextual environments that may influence emoji selection and interpretation. An emoji accompanying an Arabic segment may perform different pragmatic functions from the same emoji appearing alongside an English segment, even when the surface meaning appears similar [39].

In digital communication, CS and emoji use could operate as complementary meaning-making resources. Both phenomena contribute to the expression of stance, emotion, identity, and interpersonal alignment (cf. [15], [41]). Emojis can reinforce the communicative intentions associated with a language switch, intensify emotional content, or help clarify pragmatic meanings that emerge across languages [39]. From a machine learning perspective, this interaction creates a complex representational challenge. Put simply, monolingual models may capture language-specific emoji associations more effectively within a single linguistic system, whereas multilingual models may possess advantages in learning cross-linguistic patterns that emerge when Arabic-English code-switched within the same discourse (cf. [32]). However, DSPP may provide monolingual models like MARBERT more advantages that multilingual coverage.

### 2.3. Emojis as digital pragmatics instances

Emojis have become an integral part of digital communication, functioning as visual pragmatic and semiotic resources that complement written language [39]. In digital pragmatics, they operate as paralinguistic and contextualization cues that contribute to meaning-making beyond the verbal level, helping users express emotions, attitudes, interpersonal stances, and communicative intentions. Early studies of emoticons and emojis emphasized their roles in expressing emotion and clarifying illocutionary force [10]. More recent research has extended this perspective by treating emojis as context-dependent resources whose pragmatic interpretation emerges from their relationship with surrounding discourse (see e.g., [3]).

Emojis have emerged as a central component of online communication and the digital age, functioning as visual pragmatic/semiotic resources that complement written language in digital environments (see e.g., [39]. In digital pragmatics, emojis are widely understood as paralinguistic cues that contribute to meaning-making beyond the verbal level, helping interlocutors deliver messages successfully, express emotions, attitudes, and interpersonal stances in online interaction. Early studies on emoticons and emojis social media platforms (see e.g., [10]) emphasized their role in expressing emotion and clarifying illocutionary force. More recent research has extended this perspective, arguing that emojis should be treated not merely as emotional indicators but as contextualization cues that shape discourse interpretation and interactional alignment. With the rise of social media platforms such as X, Instagram, and Facebook, emoji use has become increasingly frequent and structurally embedded in everyday digital communication. On X in particular, the limited character space encourages users to rely on multimodal strategies, including emojis, to enhance expressivity and compress meaning.

Studies in social media linguistics have demonstrated that emojis contribute to stance-taking, identity construction, and audience engagement [39]. They are often used to reinforce opinions, soften criticism, or establish solidarity with interlocutors. In this context, emojis function as interactional resources that facilitate pragmatic efficiency in highly condensed textual environments.

### 2.4. Previous studies

Though the topic we are examining has not been studied before, to the best of our knowledge, there are some studies that are relevant to ours. We synthesize these studies into 3 themes. First, the influenced of linguistic domain-specific on the performance of PLMs. Recent research on DSPP and domain-adaptive pretraining has increasingly examined whether adapting pretrained representations to the characteristics of a target domain can improve downstream performance. [42], for example, demonstrated that incorporating domain-specific corpora into BERT-based pretraining improved performance on natural language processing tasks in the architecture, engineering, and construction domain. Similarly, research on domain-specific models has shown that pretrained representations can be tailored to specialized linguistic or professional environments, producing models whose representations are more closely aligned with the requirements of particular downstream tasks [12], [6]. Recent surveys further indicate that domain-specific adaptation has become an important strategy for improving the usefulness of pretrained language and large language models across specialized application areas [29], [14].

Additionally, Que et al. [27] examined domain-specific continual pretraining and showed that the composition of general-domain and domain-specific data can affect downstream performance, highlighting the importance of the relationship between pretraining data and target-domain requirements. This perspective suggests that pretraining should not be understood simply as a binary distinction between general-purpose and domain-specific models. Instead, models can be viewed as having different pretraining profiles, reflecting differences in language coverage, domain specialization, data composition, and the linguistic characteristics represented in their training data. Thus, a central issue is whether the degree of alignment between a model's pretraining profile and the linguistic and communicative characteristics of a target task is reflected in downstream Transformer performance.

A second strand is the relevance of pretraining alignment becomes particularly apparent in Arabic digital discourse, where dialectal variation, non-standard orthography, informal vocabulary, and social-media language create challenges that may not be adequately represented in general-purpose or broadly multilingual pretrained models. Abdul-Mageed et al. [1] addressed this issue directly through the development of MARBERT, a Transformer model pretrained on large-scale Arabic Twitter data with particular attention to dialectal Arabic and social-media language. Across a broad range of Arabic language and social-media classification tasks, MARBERT demonstrated strong performance and, on several tasks, outperformed more broadly multilingual alternatives. These findings provide important evidence that a model whose pretraining profile is closely aligned with the linguistic and communicative characteristics of Arabic social-media discourse can provide advantages for downstream classification.

Such alignment may become even more consequential in code-switched discourse, where multiple linguistic systems can occur within the same textual sequence. Arabic–English code-switching may involve changes not only in vocabulary but also in script, morphology, orthographic conventions, and patterns of informal digital expression. Research on Arabic–English code-switched data has thus examined contextual representations and multilingual Transformer approaches as a means of capturing the linguistic complexity of mixed-language digital communication. More recent work on multilingual digital texts likewise demonstrates that Transformer-based approaches can identify and model code-switching patterns in socially mediated multilingual data [23]. These studies establish code-switched digital discourse as a particularly demanding environment for language models because the relevant linguistic information may be distributed across languages and may interact with informal and socially conditioned patterns of language use.

The third and final theme is Transformer-based models can be used to computationally model interpersonal pragmatic functions of emojis in Arabic digital discourse. for example, Shormani (2026) operationalized 5 interpersonal pragmatic functions: *Politeness, Respect, Solidarity, Empathy, and Encouragement*, and finetuned MARBERT within a multi-label classification framework. The model achieved 93% accuracy, with a micro F1-score of 0.61 and a macro F1-score of 0.56, indicating that MARBERT can learn meaningful patterns associated with interpersonal communication beyond conventional sentiment or emotion classification. However, performance varied across functions: Politeness and Respect were identified more successfully than Solidarity, suggesting that computational models may handle relatively explicit and conventionalized interpersonal meanings more readily than highly context-dependent and implicit social meanings. emojis were initially studied largely in relation to emotion and affect, but research in digital pragmatics has increasingly demonstrated that they can perform a much wider range of communicative functions. This development is particularly relevant to Transformer-based approaches because contextualized representations provide a mechanism for modeling relationships between an emoji and the linguistic material surrounding it.

### 2.5. Research Gap

However, to the best of our knowledge, no prior study has systematically examined how different PLMs, representing different degrees of linguistic and domain specialization, perform in Modeling emoji pragmatic functions in Arabic-English code-switched tweets [28]. Although previous research has addressed multilingual processing, Arabic social-media language, and sentiment-related classification (e.g., [32], [43], [8]), the intersection of pretraining specialization, Arabic-English code-switching, and emoji pragmatics remains underexplored. This study addresses this gap by comparing 2 models, viz., a broadly multilingual model, XLM-R, and an Arabic social-media-specialized model, MARBERT, with BERT serving as a general-purpose baseline, thereby examining whether greater linguistic and domain alignment in pretraining is reflected in the classification of digital-pragmatic emoji instances. It thus asks:

1. Does pretraining profile influence PLMs' performance in Modeling emoji pragmatic functions in Arabic-English code-switched posts?
1. How do MARBERT and XLM-R differ in this task given their DSPP?
2. What standard metrics scores do the 2 models achieve and how do they perform in class-level modeling?

## 3. Methodology

### 3.1. Data collection and tools

We used Python 3.11.9 to collect the data from X. We collected 13397 bilingual (English and Arabic) raw data tweets, but 1594 tweets did not contain emojis, and 108 empty rows, which were eventually excluded from the corpus. Training data consist of 8000 tweets and validation and testing data 1847 and 1848, respectively (almost reflecting data splitting approach). In the preprocessed dataset, emojis were classified according to their pragmatic functions. Table 1 summarizes the 9 emoji categories found in our data (cf. [31]).

**Table 1: Emoji annotation procedure**

| Cat | Full form | Arabic meaning | Emoji examples |
|---|---|---|---|
| **HUM** | Humor | فكاهة | 😂 🤣 😅 |
| **SAR** | Sarcasm | سخرية | 😉 😎 🤔 |
| **HAP** | Happiness | سعادة | 🙂 😐 😁 |
| **LOV** | Love | حب | 😍 💞 ❤️ |
| **SAD** | Sadness | حزن | 😭 💔 🖤 |
| **AGR** | Agreement | موافقة | 👍 👌 👏 |
| **PRY** | Prayer | دعاء | 🙏 🤲 |

| PRD | Pride | فخر | 🔥 💥 💪 |
|---|---|---|---|
| FER | Fear | خوف | 😨 😦 🥶 |

These 9 categories represent the most recurrent emojis in our data. The training data contain CS tweets with emojis, each tweet contains at least one emoji. The CS nature is that each tweet contains an English expression, consisting of at least 2-word expression. Regular expressions and specialized emoji detection libraries were deployed to isolate, track, and preserve emoji tokens within the data files, which were saved in structured .xlsx (excel) sheets for downstream processing. All these processes are detailed in the following (sub)sections.

### 3.2. Pipeline

This study follows a supervised machine-learning paradigm, combining a manually annotated tweet dataset with finetuned Transformer-based pretrained language models (PLMs). All experiments were implemented in Google Colab using the Hugging Face Transformers library with a Tesla T4 GPU accelerator (cf. [34]). 2 PLMs representing distinct pretraining profiles were finetuned and evaluated under identical experimental settings, MARBERT, and XLM-R. Although used as a general-purpose baseline, BERT was also adapted to the same finetuning. The comparison focuses on whether differences in linguistic and domain specialization in pretraining are associated with differences in emoji-pragmatic classification performance in Arabic-English code-switched discourse.

#### 3.2.1. Annotation procedure

The annotation scheme is grounded in a pragmatic-functional classification of emojis, designed to capture communicative and intent-based aspects of interaction (cf. [31]). Emojis are analyzed as context-sensitive signals of speaker intention but not isolated lexical units, signifying mainly sentiments or emotions (cf. [30]). The scheme comprises 9 functional categories. Manual annotation was conducted by 3 trained bilingual annotators following a predefined codebook. To ensure annotation quality, the annotators underwent a 2-week training course consisting of 4 hours of guided review per day. Annotators followed the following procedure:

1. Read the tweet in its full, un-normalized contextual form to understand the communicative context.
2. Identify the target emoji(s) and determine their communicative role within the code-switched tweet.
3. Assign a single, dominant pragmatic label from the 9-class codebook (cf. Table 1).
4. Base the decision primarily on contextual meaning and pragmatic interpretation but not literal or purely lexical meaning.
5. For tweets containing multiple emojis, assign the most salient pragmatic function.

In cases of disagreement, majority voting and adjudication were used to establish the final gold-standard labels.

#### 3.2.2. Data preprocessing and tokenization

All tweets were processed using Pandas to handle missing fields and maintain consistent UTF-8 text encoding. Emojis were preserved in their original textual form without normalization or conversion into textual descriptions. Pragmatic labels were converted into numerical identifiers using a predefined label-to-index mapping. The annotated corpus was divided into 3 predefined subsets: 8000 tweets for training, 1847 for validation, and 1848 for independent testing. Stratified sampling (random seed = 42) was used to preserve the proportional distribution of the 9 pragmatic classes across the 3 subsets.

Each dataset was tokenized using the tokenizer associated with the respective pretrained model, with truncation and padding applied to a maximum sequence length of 128 tokens. Given that the models differ in their pretraining profiles and tokenization resources, their tokenization mechanisms are relevant to understanding their suitability for the target data. BERT uses an English-oriented WordPiece vocabulary and was not specifically pretrained for Arabic social-media language; consequently, Arabic forms may be represented less efficiently or mapped to the UNK. MARBERT uses an Arabic-focused tokenizer associated with pretraining on large-scale Arabic social-media data, making it more closely aligned with dialectal and informal Arabic. Its social-media pretraining may also provide exposure to English vocabulary and script occurring in such data. XLM-R uses a SentencePiece-based multilingual tokenizer and was pretrained on more than 100 languages, including Arabic, providing broad multilingual coverage for mixed-language input [25]. These differences are considered as characteristics of the models' respective pretraining profiles but not as independently manipulated variables.

### 3.2.3. Optimization and finetuning

The finetuning pipeline used supervised learning with the PyTorch-backed HuggingFace Trainer. Hyperparameters were held constant across all 3 models to ensure a controlled comparison (cf. [13], [17]). The models were finetuned using the AdamW optimizer with a batch size of 16 for both training and evaluation, a learning rate of $2 \times 10^{-5}$, and a weight decay of 0.01. Training was conducted for 5 epochs, with validation performed after each epoch. Mixed-precision FP16 was enabled during GPU training, and the maximum sequence length was set to 128 tokens. For model selection, the checkpoint achieving the highest validation Macro-F1 was retained for independent testing. Accuracy, precision, recall, Macro-F1, Weighted F1 and validation loss were recorded at each epoch to monitor model performance throughout training.

A linear sequence-classification head was added to each pretrained model for the 9-class pragmatic classification task. The comparative analysis included 3 distinct pretraining profiles: i) BERT, initialized from *bert-base-uncased*, representing general-purpose English pretraining and serving as a baseline for processing Arabic-English code-switched data [9], ii) MARBERT, initialized from *UBC-NLP/MARBERT*, representing Arabic-focused social-media pretraining based on approximately one billion Arabic tweets and providing strong linguistic and domain alignment with the target data [1], [11], and iii) XLM-R , initialized from *XLM-R -base*, representing broad multilingual pretraining on large-scale multilingual data and providing cross-lingual representations across more than 100 languages [7]. All 3 models were trained under the same conditions, allowing their performance differences to be interpreted in relation to their differing pretraining profiles.

### 3.2.4. Methods of analysis

This study adopts a quantitative and qualitative NLP approach to investigate the role of 2 Transformer-based PLMs representing different pretraining profiles. It conceptualizes emojis as context-sensitive, paralinguistic pragmatic markers that contribute to meaning construction in digital communication (cf. [16]). Mathematically and due to the uneven distribution across the 9 emoji pragmatic categories, evaluating performance based on accuracy alone could mask class-specific errors. Accordingly, model evaluation relies on 5 standard metrics: Accuracy, Macro Precision, Macro Recall, Macro F1-score, and Weighted F1-score (see also [18]). The performance metrics employed in this study are formally defined as follows.

$$Accuracy = \frac{TP + TN}{TP + TN + FP + FN}$$

$$Macro\ Precision = \frac{1}{|Y|}\sum_{c \in Y}\frac{TP_c}{TP_c + FP_c}$$

$$Macro\ Recall = \frac{1}{|Y|}\sum_{c \in Y}\frac{TP_c}{TP_c + FN_c}$$

$$Macro\ F1 - Score = \frac{1}{|Y|}\sum_{c \in Y}\frac{2 \cdot Precision_c \cdot Recall_c}{Precision_c + Recall_c}$$

$$Weighted\ F1 - Score = \sum_{c \in Y}\frac{n_c}{N}F1_c$$

Macro-averaged Precision, Recall, and F1-score are calculated independently for each class and then averaged (e.g., Fear and Pride) contribute equally to the overall evaluation. In addition, Weighted F1-score is additionally reported by weighting each class according to its support, thereby accounting for the uneven distribution of the pragmatic categories. In addition to numerical evaluation, confusion matrices and training dynamics (training loss (TL), validation loss (VL), and Macro-F1 trajectories were analyzed to examine model convergence, class-level prediction behavior, and optimization stability. This task is formulated as a multi-class sequence classification problem. Let $D = \{(x_i, y_i)\}^{N}{}_{i=1}$ be a dataset consisting of $N$ tweets, where each $x_i$ input consists of an arbitrary mixture of Arabic and English tokens. We define a mapping function $f: x \rightarrow y$, where $y = \{HUM, SAR, HAP, LOV, SAD, AGR, PRY, PRD, FER\}$ denotes the set of 9 distinct emoji classes.

## 4. Results

In this section, we present the study results, tabulating and analyzing them. To start with, Tables 2-4 summarize the 3 models' results.

**Table 2: MARBERT train_valid results**

| Epoch | Training Loss | Validation Loss | Accuracy | Precision | Recall | Macro F1 | Weighted F1 |
|---|---|---|---|---|---|---|---|
| **1** | 0.877000 | 0.551548 | 0.866412 | 0.368424 | 0.424601 | 0.393825 | 0.053151 |
| **2** | 0.396300 | 0.285184 | 0.923664 | 0.570083 | 0.593761 | 0.575761 | 0.093336 |
| **3** | 0.145700 | 0.203440 | 0.954198 | 0.689574 | 0.746970 | 0.714097 | 0.145198 |
| 4 | 0.093300 | 0.197260 | 0.958015 | 0.827526 | 0.860202 | 0.842640 | 0.696452 |
| 5 | 0.053100 | 0.185813 | 0.958015 | 0.827526 | 0.860202 | 0.842640 | 0.875234 |

**Table 3: XLM-R train_valid results**

| Epoch | Training Loss | Validation Loss | Accuracy | Precision | Recall | Macro F1 | Weighted F1 |
|---|---|---|---|---|---|---|---|
| **1** | 1.587200 | 1.505636 | 0.480916 | 0.053435 | 0.111111 | 0.072165 | 0.253693 |
| **2** | 1.148400 | 0.638751 | 0.793893 | 0.246803 | 0.321321 | 0.276469 | 0.194381 |
| **3** | 0.423400 | 0.357880 | 0.885496 | 0.370432 | 0.430806 | 0.393996 | 0.351057 |
| 4 | 0.311500 | 0.268889 | 0.893130 | 0.376903 | 0.436812 | 0.398695 | 0.511452 |
| 5 | 0.240700 | 0.232316 | 0.923664 | 0.522144 | 0.545145 | 0.512869 | 0.745325 |

**Table 4: BERT train_valid results**

| Epoch | Training Loss | Validation Loss | Accuracy | Precision | Recall | Macro F1 | Weighted F1 |
|---|---|---|---|---|---|---|---|
| **1** | 1.361743 | 1.517963 | 0.480916 | 0.053435 | 0.111111 | 0.072165 | 0.567412 |
| **2** | 1.584627 | 1.478807 | 0.480916 | 0.053435 | 0.111111 | 0.072165 | 0.584634 |
| **3** | 1.426214 | 1.435047 | 0.519084 | 0.168170 | 0.149200 | 0.134543 | 0.426359 |
| 4 | 1.381319 | 1.397742 | 0.549618 | 0.199059 | 0.180238 | 0.174844 | 0.381543 |
| 5 | 1.325632 | 1.387856 | 0.549618 | 0.205215 | 0.180238 | 0.175732 | 0.323631 |

**Table 5: Model performance on the Test Set**

| Model | Accuracy | Precision | Recall | Macro F1 | Weighted F1 |
|---|---|---|---|---|---|
| **MARBERT** | 0.96 | 0.83217 | 0.87120 | 0.853150 | 0.875210 |
| **XLM-R** | 0.92 | 0.52219 | 0.52735 | 0.522869 | 0.737196 |

As Table 2 displays, MARBERT achieved the strongest overall performance across all evaluation metrics, outperforming XLM-R. During its 5-epoch training run, MARBERT exhibited rapid convergence, steadily reducing its validation loss from 0.55 to 0.19 while substantially improving its Macro F1 score from 0.39 to 0.84. On the independent test set (Table 5), MARBERT maintained its top position, achieving an accuracy of 0.96, precision of 0.83, recall of 0.87, a Macro F1 score of 0.85, and a Weighted F1 score of 0.88

Table 3 shows that XLM-R demonstrated stable learning dynamics throughout training with VL dropping from 1.51 in the first epoch to 0.23 the fifth epoch. Its validation Macro F1 score improved from 0.07 to 0.51, while its Weighted F1 reached 0.75. On the test set (Table 5), XLM-R achieved competitive secondary results, securing an accuracy of 0.92, precision of 0.52, recall of 0.53, Macro F1 of 0.52 and Weighted F1 of 0.74. While showing effective generalizability, it remained significantly behind MARBERT across all metric dimensions.

By way of comparison, BERT struggled to adapt effectively within the 5-epoch limit (Table 4). Its VL remained elevated throughout training, decreasing only slightly from 1.52 to 1.39. BERT yielded limited class-discriminative ability, capping its validation performance at epoch 5 at an accuracy of 0.55, precision of 0.21, recall of 0.18, and a Macro F1 score of 0.18. These results underscore the substantial advantage of DSPP, MARBERT and robust multilingual capabilities XLM-R over standard English-centric pretraining on Arabic-English CS data. These are also supported by Figs 1 & 2.

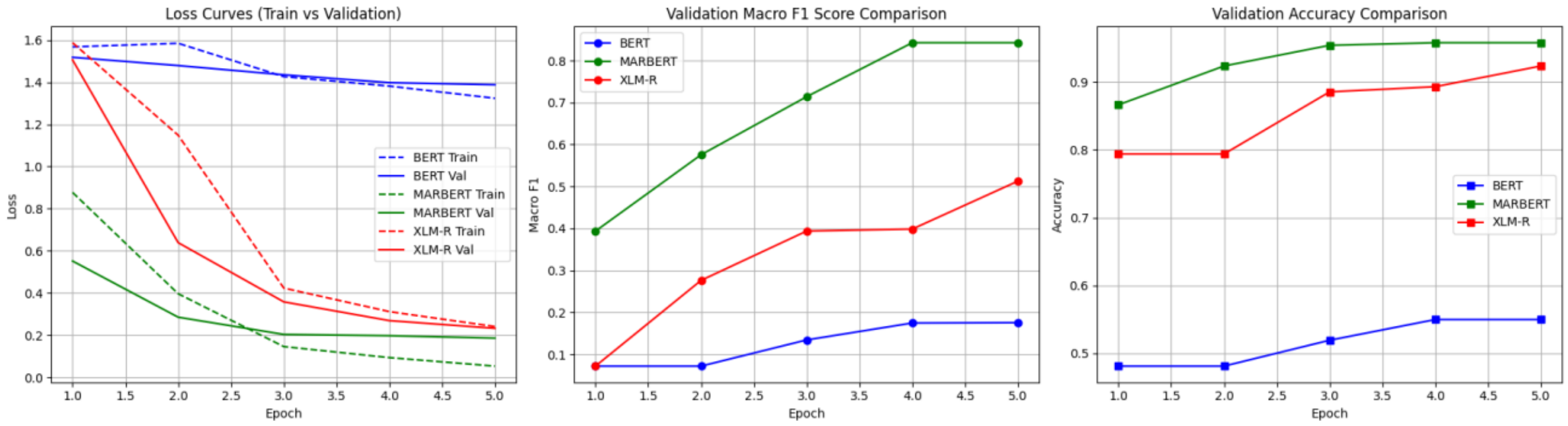

***Fig1: Learning curves***

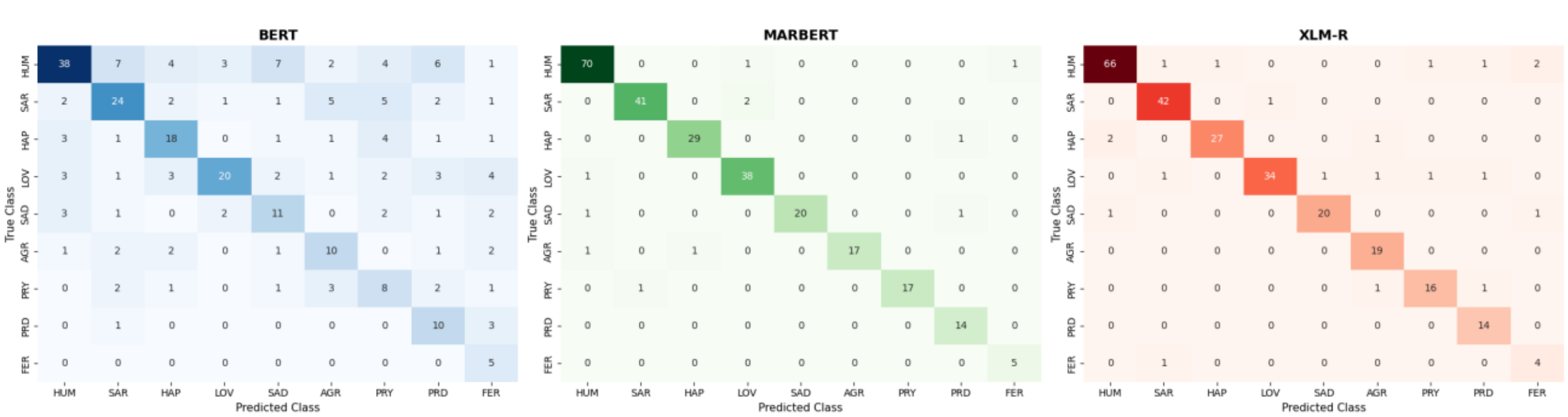

***Fig 3: Confusion matrices***

Figure 3 details the class-level classification behavior across models using validation confusion matrices at Epoch 5. MARBERT achieved sharp, dominant diagonal concentration, indicating high classification precision and minimal inter-class confusion across all categories, including majority classes such as HUM (70 true positives) and SAR (41 true positives). XLM-R also captured a distinct diagonal pattern with strong performance on major classes (66 for HUM, 42 for SAR), though minor off-diagonal leakage occurred across several lower-frequency classes, explaining its reduced Macro F1 score relative to MARBERT. BERT showed heavy off-diagonal dispersion across nearly all class boundaries, demonstrating significant misclassification rates and a strong bias toward majority classes, confirming its inability to effectively learn discriminative feature boundaries for this task.

## 5. Discussion

The findings provide clear evidence that DSPP is critical to the classification of pragmatic functions in Arabic–English code-switched digital discourse. Under identical finetuning conditions, the 3 models exhibited markedly different learning trajectories and levels of discriminative performance. MARBERT consistently achieved the strongest results, XLM-R occupied an intermediate position, and the general-domain BERT baseline showed substantially weaker learning. This performance ordering is particularly informative because the task involves short, informal, socially situated utterances in which pragmatic interpretation depends not only on lexical content but also on dialectal variation, code-switching, discourse context, and the multifunctionality of pragmatic expressions.

MARBERT outperformed XLM-R precisely due to its advantage of domain-aligned pretraining, demonstrating the most effective adaptation to the classification task. Its VL decreased from 0.55 in epoch 1 to 0.19 in epoch 5, while validation Macro F1 increased from 0.39 to 0.84—an absolute improvement of 0.45. The simultaneous reduction in VL and increase in Macro F1 indicates that the model was not merely improving overall classification accuracy, but was progressively developing stronger discrimination across the 9 pragmatic-function categories. Its validation accuracy increased from 0.87 to 0.96, while macro precision increased from 0.37 to 0.83 and macro recall from 0.42 to 0.86. The independent test results further support this pattern: MARBERT achieved an accuracy of 0.96, macro precision of 0.83, macro recall of 0.87, and Macro F1 of 0.85. The close correspondence between its final validation Macro F1 (0.84) and independent test Macro F1 (0.85) is particularly important. The model maintained and slightly improved its Macro F1 performance, suggesting that its learned representations generalized effectively beyond the examples used during training.

The strong performance of MARBERT is theoretically plausible given its pretraining profile X' posts. MARBERT was specifically pretrained on a very large collection of Arabic tweets, encompassing dialectal Arabic and Modern Standard Arabic across approximately one billion Arabic tweets. Its pretraining corpus was therefore directly aligned with the linguistic environment represented in this task: informal social-media language, dialectal Arabic, short textual units, and naturally occurring code-switching. This alignment is especially relevant to digital pragmatics. Pragmatic functions—such as humor, sarcasm, agreement, prayer, affection, sadness, and fear—cannot always be identified through isolated lexical items. Their interpretation depends heavily on informal constructions, dialectal wording, code-switching, discourse conventions, and contextual cues. Exposure to vast quantities of naturally occurring social-media discourse equips MARBERT with rich representations specifically tailored to these conditions. MARBERT's validation Macro F1 peaks at 0.84 in both epochs 4 and 5, while its weighted F1 increases from 0.70 to 0.88.

That means the model's final epoch was not selected because Macro F1 improved—it remained exactly the same at 0.84—but the weighted F1 increased substantially. If your code says that the best checkpoint is selected according to validation Macro F1, then epoch 4 and epoch 5 are tied on the selection metric. This should be handled consistently in the methodology/code and, if relevant, reported transparently.

However, XLM-R demonstrated strong general learning but weaker pragmatic specialization. It showed substantial learning during finetuning, although its final performance remained well below that of MARBERT. Its VL declined sharply from 1.51 in epoch 1 to 0.23 in epoch 5, while validation Macro F1 increased from 0.07 to 0.51 (an absolute Macro F1 improvement of 0.44). Validation accuracy increased from 0.48 to 0.92, with macro precision and recall reaching 0.52 and 0.55, respectively, by epoch 5. Independent test results confirm that XLM-R learned useful task representations, achieving 0.92 accuracy, 0.52 macro precision, 0.53 macro recall, and 0.52 Macro F1. However, its test Macro F1 was 0.33 lower than MARBERT's (0.85 versus 0.52). Thus, while XLM-R generalized substantially beyond chance level, it lacked the balanced, fine-grained class discrimination achieved by domain-specialized MARBERT.

The superior performance of MARBERT is also support by class-level performance. Table 6 summarizes this.

**Table 6: Class-level performance**

| Class | Support | MARBERT | XLM-R | Δ F1 |
|---|---|---|---|---|

| | | P | R | F1 | P | R | F` | |
|---|---|---|---|---|---|---|---|---|
| **HUM** | 72 | 0.96 | 0.97 | 0.97 | 0.76 | 0.53 | 0.62 | +0.34 |
| **SAR** | 43 | 0.98 | 0.95 | 0.96 | 0.62 | 0.56 | 0.59 | +0.38 |
| **HAP** | 30 | 0.97 | 0.97 | 0.97 | 0.58 | 0.60 | 0.59 | +0.38 |
| **LOV** | 39 | 0.93 | 0.97 | 0.95 | 0.77 | 0.51 | 0.62 | +0.33 |
| **SAD** | 22 | 1.00 | 0.91 | 0.95 | 0.50 | 0.50 | 0.50 | +0.45 |
| **AGR** | 19 | 1.00 | 0.89 | 0.94 | 0.45 | 0.53 | 0.49 | +0.46 |
| **PRY** | 18 | 1.00 | 0.94 | 0.97 | 0.32 | 0.44 | 0.37 | +0.60 |
| **PRD** | 14 | 0.88 | 1.00 | 0.93 | 0.40 | 0.71 | 0.51 | +0.42 |
| **FER** | 5 | 0.83 | 1.00 | 0.91 | 0.25 | 1.00 | 0.40 | +0.51 |

*Note: P = precision, R = recall, F1 = F1 score, Δ F1 = difference in F1 score between MARBERT and XLM-R*

Table 6 presents MARBERT and XLM-R's performance on the test set. The class-level results provide important evidence concerning the nature of the performance difference between MARBERT and XLM-R. MARBERT outperformed XLM-R in all 9 pragmatic-function categories, indicating that its superior overall Macro F1 was not attributable to a small number of particularly well-classified categories. The F1 advantage ranged from 0.33 for love (LOV) to 0.60 for prayer (PRY), with substantial differences also observed for fear (FER; +0.51), agreement (AGR; +0.46), sadness (SAD; +0.45), and praise (PRD; +0.42). This systematic pattern complements the aggregate results reported above and demonstrates that MARBERT provided more balanced discrimination across the pragmatic-function inventory.

The precision–recall profiles further illustrate the difference between the two models. MARBERT achieved recall values of at least 0.89 across all 9 categories, whereas XLM-R showed substantially lower recall for most categories. The difference was particularly pronounced for humor, love, and prayer. The fear category provides an additional illustration: both models achieved perfect recall, but MARBERT obtained substantially higher precision (0.83 versus 0.25), indicating that XLM-R generated many more false-positive fear predictions. Thus, MARBERT's advantage was reflected not simply in recovering more instances of the target categories but also in making more precise category assignments.

This consistent class-level superiority is compatible with the proposed role of DSPP in Arabic social-media pragmatic classification. The task involves informal Arabic, dialectal variation, code-switching, and pragmatic meanings that may depend on socially situated and context-sensitive language use. MARBERT's pretraining on Arabic social-media language provides a theoretically relevant source of linguistic and contextual specialization for such a task. The fact that MARBERT outperformed XLM-R across all 9 pragmatic categories is thus consistent with the possibility that domain-aligned representations contributed to its stronger performance. However, our experiment provides empirical support for this explanation rather than as definitive causal evidence.

Recall that XLM-R was designed as a multilingual model trained on a very large multilingual corpus, but not specifically optimized for NAD social-media texts and discourse. Its multilingual representation provides an important advantage for code-switched data because the model is not restricted to Arabic and can represent multiple languages within a shared framework. At the same time, this broader coverage may provide less specialized exposure to the dialectal, informal, and pragmatic characteristics of Arabic social-media communication than MARBERT. These results suggest that multilingual coverage alone does not guarantee optimal performance on a highly specialized pragmatic classification task. XLM-R clearly learned the task, but its lower Macro F1 indicates that broad multilingual pretraining did not provide the same degree of class-sensitive representation as Arabic social-media-oriented pretraining. This interpretation is consistent with previous work showing the value of Arabic-specific and social-media-oriented pretraining for Arabic NLP tasks (cf. [32]).

As a general-domain baseline, BERT provides an important baseline for interpreting the performance of the specialized models. We included BERT in the fine-tuning and validation pipeline under identical experimental conditions and hyperparameter settings as MARBERT and XLM-R. It showed considerably weaker adaptation under the same 5-epoch finetuning regime. Its VL decreased only from 1.52 in epoch 1 to 1.39 in epoch 5, a reduction of just 0.13. During the same period, validation accuracy increased from 0.48 to 0.55, while Macro F1 increased from 0.07 to only 0.18. The final validation Macro F1 was 0.67 points lower than MARBERT's final validation Macro F1 (0.84) and 0.34 points lower than XLM-R's (0.51). The difference is particularly striking because all the 3 models were finetuned under the same basic experimental conditions. While this performance gap is striking, it aligns with expectations given BERT's pre-training footprint. Including BERT establishes an empirical lower bound, demonstrating how a general English-centric Transformer adapts to digital pragmatics in code-switched text without exposure to NAD, social-media discourse, or multilingual specialization. Given that BERT was excluded from final test-set evaluation due to its poor convergence, this comparative baseline relies strictly on the validation dynamics reported in Table 4.

## 6. Conclusions and limitations

To conclude, a number of conclusions can be drawn: i) MARBERT demonstrated the strongest and most consistent performance. Its validation Macro F1 increased from 0.39 to 0.84, while VL decreased from 0.55 to 0.19. On the independent test set, it achieved 0.96 accuracy, 0.83 macro precision, 0.87 macro recall, and 0.85 Macro F1, indicating effective generalization. MARBERT outperformed XLM-R across all 9 pragmatic-function categories, supporting the robustness of its overall performance, ii) domain alignment and multilingual coverage contribute differently to Transformer performance. Although XLM-R was trained on multilingual data, it achieved 0.92 test accuracy, its Macro F1 was substantially lower (0.52) than MARBERT's 0.85. Thus, multilingual pretraining may not provide the same advantage as domain-aligned pretraining for Arabic–English digital pragmatic classification, and iii) class-level evaluation is important for assessing Transformer performance. The differences across accuracy, Macro F1, precision, and recall demonstrate that accuracy alone does not fully capture performance across pragmatic categories.

This study suggests 3 actionable strategies: i) considering domain alignment when selecting pretrained models. Models pretrained on data that reflect the linguistic and communicative characteristics of the target domain may offer practical advantages for dialectal, informal, and code-switched discourse, ii) using class-sensitive evaluation. Macro F1, macro precision, and macro recall should complement accuracy when pragmatic categories vary in frequency (see also [30]), and iii) develop context-sensitive and explainable models. XAI techniques can help determine whether Transformer predictions reflect meaningful pragmatic signals or superficial statistical patterns.

However, this study has several limitations including: i) the dataset is restricted to Arabic–English code-switching, limiting generalization to trilingual and polyglot digital communities, ii) although the 9-class human-annotation scheme covers major pragmatic functions, some fine-grained distinctions, such as irony versus sarcasm or different forms of pride, are necessarily collapsed, potentially limiting the classification of more nuanced pragmatic meanings, and iii) the models were trained with a fixed 128-token sequence length, which may constrain the representation of longer threads and multi-turn interactions. Future work will address these limitations through attention visualization and other XAI approaches to examine Transformer decision-making and through parameter-efficient finetuning of larger models for multilingual pragmatic-function classification.